\documentclass[10pt,twocolumn,letterpaper]{article}

\usepackage{cvpr}
\usepackage{times}
\usepackage{epsfig}
\usepackage{graphicx}
\usepackage{amsmath}
\usepackage{amssymb}

\usepackage{subfigure}
\usepackage{mathrsfs}
\usepackage{amsthm}
\usepackage{algorithm}
\usepackage{algorithmic}
\usepackage{enumerate}
\usepackage{multirow}

\usepackage{enumitem}
\usepackage{makecell}
\usepackage{tabulary}
\usepackage{pifont}

\usepackage{boldline}

\usepackage[breaklinks=true,bookmarks=false]{hyperref}

\cvprfinalcopy 

\def\cvprPaperID{5422} 

\ifcvprfinal\fi
\begin{document}

\title{Exploring Object Relation in Mean Teacher for Cross-Domain Detection\thanks{{\small This work was performed at JD AI Research.}}}

\author{Qi Cai $^{\dag}$, Yingwei Pan $^{\ddag}$, Chong-Wah Ngo $^{\S}$, Xinmei Tian $^{\dag}$, Lingyu Duan $^{\P}$, and Ting Yao $^{\ddag}$ \\
{\small\centering$^{\dag}$ University of Science and Technology of China, Hefei, China}~~~~~
{\small\centering$^{\ddag}$ JD AI Research, Beijing, China}\\
{\small\centering$^{\S}$ City University of Hong Kong, Kowloon, Hong Kong}~~~~~
{\small\centering$^{\P}$ Peking University, Beijing, China}\\
{\tt\scriptsize \{cqcaiqi, panyw.ustc, tingyao.ustc\}@gmail.com, cscwngo@cityu.edu.hk, xinmei@ustc.edu.cn, lingyu@pku.edu.cn}
}

\input{def.set}
\maketitle
\thispagestyle{empty}

\begin{abstract}
	Rendering synthetic data (e.g., 3D CAD-rendered images) to generate annotations for learning deep models in vision tasks has attracted increasing attention in recent years. However, simply applying the models learnt on synthetic images may lead to high generalization error on real images due to domain shift. To address this issue, recent progress in cross-domain recognition has featured the Mean Teacher, which directly simulates unsupervised domain adaptation as semi-supervised learning. The domain gap is thus naturally bridged with consistency regularization in a teacher-student scheme. In this work, we advance this Mean Teacher paradigm to be applicable for cross-domain detection. Specifically, we present Mean Teacher with Object Relations (MTOR) that novelly remolds Mean Teacher under the backbone of Faster R-CNN by integrating the object relations into the measure of consistency cost between teacher and student modules. Technically, MTOR firstly learns relational graphs that capture similarities between pairs of regions for teacher and student respectively. The whole architecture is then optimized with three consistency regularizations: 1) region-level consistency to align the region-level predictions between teacher and student, 2) inter-graph consistency for matching the graph structures between teacher and student, and 3) intra-graph consistency to enhance the similarity between regions of same class within the graph of student. Extensive experiments are conducted on the transfers across Cityscapes, Foggy Cityscapes, and SIM10k, and superior results are reported when comparing to state-of-the-art approaches. More remarkably, we obtain a new record of single model: 22.8\% of mAP on \textit{Syn2Real} detection dataset.
\end{abstract}

\section{Introduction}
\begin{figure}[!tb]
	\vspace{-0.02in}
	\centering {\includegraphics[width=0.36\textwidth]{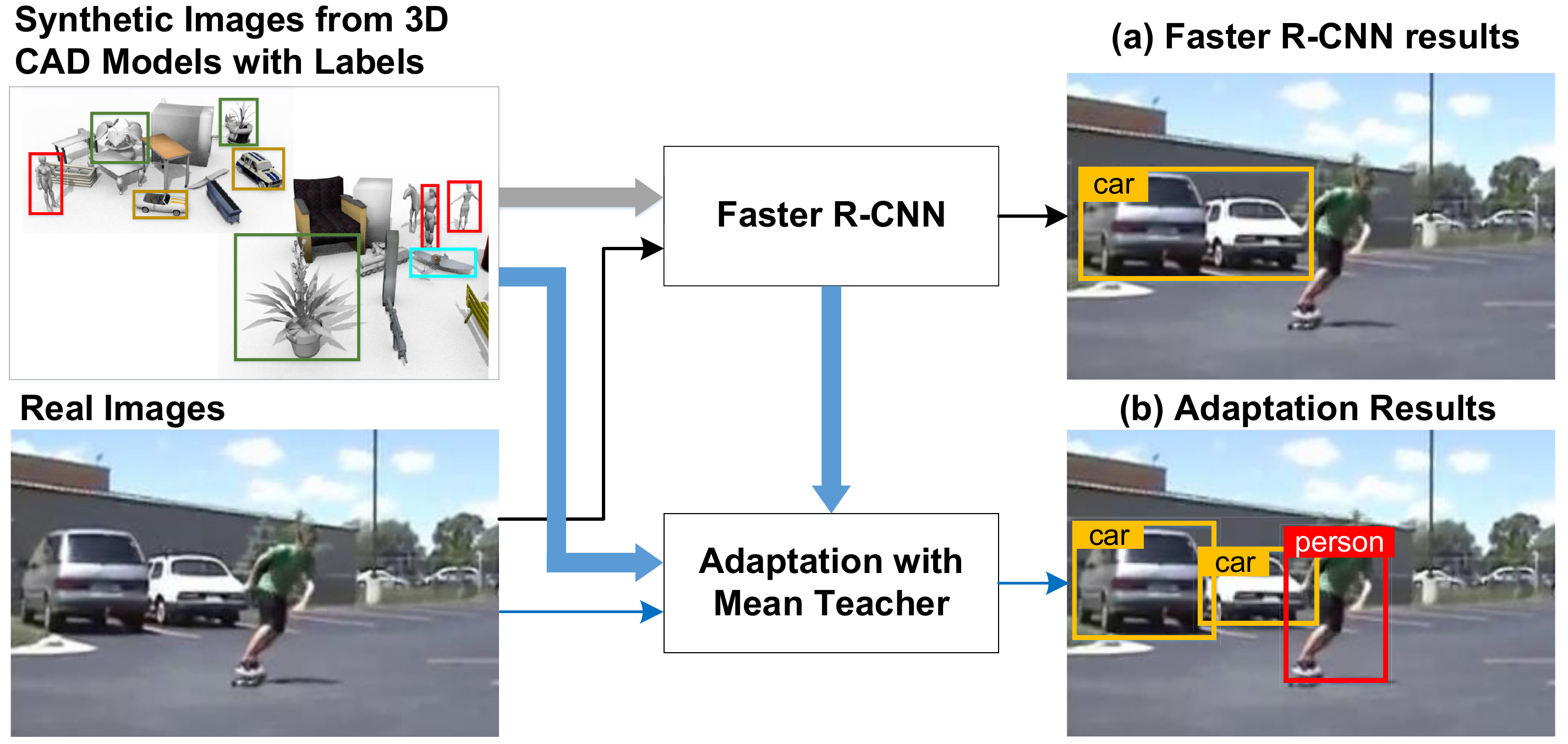}}
	\vspace{-0.05in}
	\caption{\small Object detection on one real image by (a) directly applying Faster R-CNN trained on images from 3D CAD models and (b) domain adaptation of Mean Teacher in this work.}
	\label{intro1}
	\vspace{-0.30in}
\end{figure}

Deep Neural Networks have been proven to be highly effective for learning vision models on large-scale datasets. To date in the literature, there are various datasets (e.g., ImageNet \cite{ILSVRC15} and COCO \cite{lin2014microsoft}) that include well-annotated images useful for developing deep models across a variety of vision tasks, e.g., recognition \cite{he2016resnet,Szegedy:CVPR15}, detection \cite{girshick2015fast,ren2015faster}, and semantic segmentation \cite{chen2018deeplab,long2015fully}. Nevertheless, given a new dataset, the typical first step is still to perform intensive manual labeling, which is cost expensive and time consuming. An alternative is to utilize synthetic data which is largely available from 3D CAD models \cite{peng2018syn2real}, and the ground truth could be freely and automatically generated. However, many previous experiences have also shown that reapplying a model learnt on synthetic data may hurt the performance on real data due to a phenomenon known as ``domain shift" \cite{yao2012predicting}. Take the object detection results shown in Figure \ref{intro1} (a) as an example, the model trained on synthetic data from 3D CAD fails to accurately localize the objects such as person and car. As a result, unsupervised domain adaptation, which aims to utilize labeled examples from the source domain and numerous unlabeled examples in the target domain to reduce the prediction error on the target data, can be a feasible solution for this challenge.

\begin{figure}[!tb]
	\vspace{-0.0in}
	\centering {\includegraphics[width=0.48\textwidth]{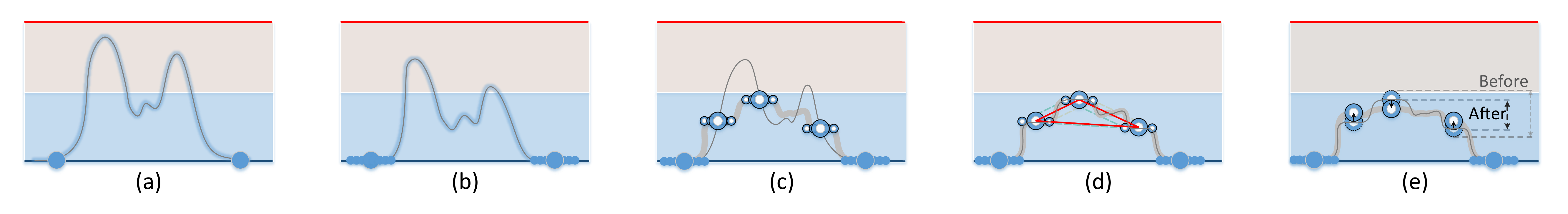}}
	\vspace{-0.25in}
	\caption{\small A sketch of cross-domain binary classification task with two labeled examples/regions in source domain (large blue dots) and three unlabeled examples/regions of one image in target domain (blue circle), demonstrating how the choice of the unlabeled target samples affects the unified fitted function across domains (gray curve). (a) A model with no regularization is flexible to fit any function that correctly classifies only labeled source data. (b) A model trained with augmented labeled source data (small blue dots) learns to produce consistent results around labeled data. (c) Mean Teacher \cite{french2018self} locally enforces the predictions to be consistent to the noise around each individual target sample, pursuing additional local smoothing of fitted function (gray curve). (d) Mean Teacher with inter-graph consistency simultaneously adapts target samples to make the holistic graph structure of them resistant to the noise. (e) Mean Teacher with intra-graph consistency enforces additional consistency across target samples of same class, further improving fitted function with long-range smoothing.}
	\label{intro2}
	\vspace{-0.23in}
\end{figure}

A recent pioneering practice \cite{french2018self} in unsupervised domain adaptation is to directly simulate this task as semi-supervised learning. The basic idea is to develop Mean Teacher \cite{tarvainen2017mean}, the state-of-the-art technique in semi-supervised learning, to work in cross-domain recognition task by pursuing the consistency of two predictions under perturbations of inputs (e.g., different augmentations of image). As such, the domain gap is naturally bridged via the consistency regularization in Mean Teacher, which enforces the predictions of two models (i.e., teacher and student) to be consistent to the perturbations/noise around each unlabeled target sample (Figure \ref{intro2} (c)). Mean teacher aims for learning a more smooth domain-invariant function than the model trained with no regularization (Figure \ref{intro2} (a)) or only augmented labeled source data (Figure \ref{intro2} (b)). In this paper, we novelly consider the use of Mean Teacher for cross-domain detection from the viewpoint of both region-level and graph-structured consistencies. The objective of region-level consistency is to align the region-level classification results of teacher and student models for the identical teacher-generated region proposals, which in turn implicitly enforces the consistency of object localization. The inspiration of graph-structured consistency is from the rationale that the inherent relations between objects within one image should be invariant to different image augmentations. In the context of Mean Teacher, this kind of graph-structured consistency (i.e., inter-graph consistency) is equivalent to matching the graph structures between teacher and student models (Figure \ref{intro2} (d)). Another kind of graph-structured consistency, i.e., intra-graph consistency, is additionally exploited to reinforce the similarity between image regions of same class within the graph of student model (Figure \ref{intro2} (e)).

\begin{figure*}[!tb]
	\vspace{-0.1in}
	\centering {\includegraphics[width=0.9\textwidth]{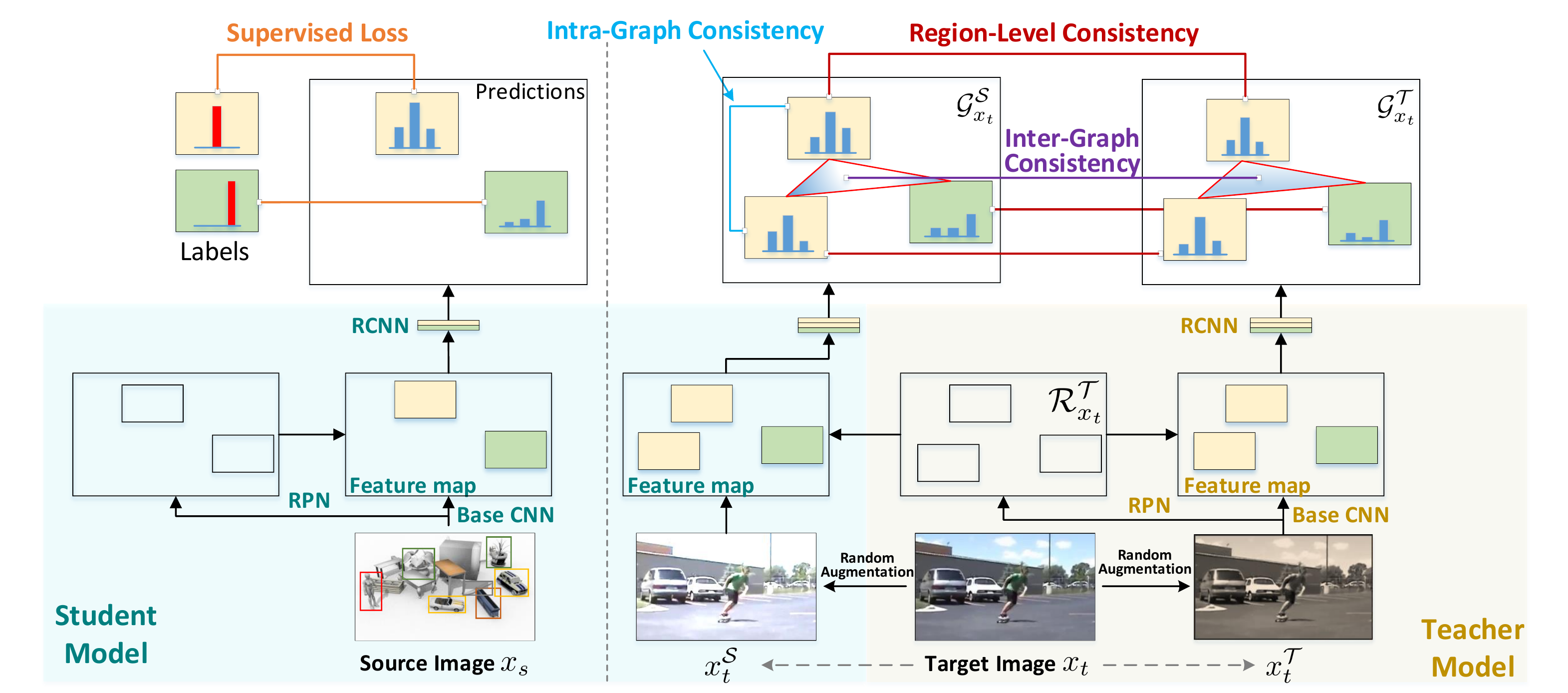}}
	\vspace{-0.05in}
	\caption{\small The overview of Mean Teacher with Object Relations (MTOR) for cross-domain detection, with teacher and student models under the same backbone of Faster R-CNN (better viewed in color). Each labeled source image is fed into student model to conduct the supervised learning of detection. Each unlabeled target image $x_t$ is firstly transformed into two perturbed samples, i.e., $x_t^{\mathcal{S}}$ and $x_t^{\mathcal{T}}$, with different augmentations and then we inject the two perturbed samples into student and teacher model separately. During training, with the same set of teacher-generated region proposals $\mathcal{R}_{x_t}^{\mathcal{T}}$ that shares between teacher and student, two relational graphs, i.e., $\mathcal{G}^{\mathcal{T}}_{x_t}$ and $\mathcal{G}^{\mathcal{S}}_{x_t}$, are constructed via calculating the feature similarity between each pair of regions for teacher and student, respectively. Next, three consistency regularization are devised to facilitate cross-domain detection in Mean Teacher paradigm from region-level and graph-structured perspectives: 1) \emph{Region-Level Consistency} to align the region-level predictions between teacher and student; 2) \emph{Inter-Graph consistency} for matching the graph structures between teacher and student, and 3) \emph{Intra-Graph Consistency} to enhance the similarity between regions of same class within the graph of student. The whole MTOR is trained by minimizing the supervised loss on labeled source data plus the three consistency losses on unlabeled target data in an end-to-end manner. Note that the student model is optimized with stochastic gradient descent and the weights of teacher are the exponential moving average of student model weights.}
	\label{fig:Framework}
	\vspace{-0.2in}
\end{figure*}

By consolidating the idea of region-level and graph-structured consistencies into Mean Teacher for facilitating cross-domain detection, we present a novel Mean Teacher
with Object Relations (MTOR), as shown in Figure \ref{fig:Framework}. The whole framework consists of teacher and student modules under the same backbone of Faster R-CNN \cite{ren2015faster}. Specifically, each labeled source sample is only passed through student module to conduct supervised learning of detection, while each unlabeled target sample will be fed into both teacher and student with two random augmentations, enabling the measure of the consistency between them to the induced noise. During training, with the same region proposals generated by teacher, two relational graphs~are~constructed via calculating the feature similarity between each pair of regions for teacher and student. The whole MTOR is then trained by the supervised detection loss in student model plus three consistency regularizations, i.e., region-level consistency to align the region-level predictions, inter-graph consistency to match the graph structures between teacher and student, and intra-graph consistency to enhance the similarity between regions of same class in student. With both region-level and graph-structured consistencies, our MTOR could better build invariance across domains and thus obtain encouraging detection results in~Figure~\ref{intro1}~(b).

\section{Related Work}
\textbf{Object Detection.} Recent years have witnessed remarkable progress in object detection with deep learning. R-CNN \cite{girshick2014rich} is one of the early works that exploits a two-stage paradigm for object detection by firstly generating region proposals with selective search and then classifying the proposals into foreground classes/background. Later Fast R-CNN \cite{girshick2015fast} extends such paradigm by sharing convolution features across region proposals to significantly speed up the detection process. Faster R-CNN \cite{ren2015faster} advances Fast R-CNN by replacing selective search with an accurate and efficient Region Proposal Networks (RPN). Next, a few subsequent works \cite{dai2016r,dai2017deformable,hu2018relation,li2017light,lin2017feature,peng2018megdet,singh2018r} strive to improve the accuracy and speed of two-stage detectors. Another line of works builds detectors in one-stage manner by skipping region proposal stage. YOLO \cite{redmon2016you} jointly predicts bounding boxes and confidences of multiple categories as regression problem. SSD \cite{liu2016ssd} further improves it by utilizing multiple feature maps at different scales. Numerous extensions to the one-stage scheme have been proposed, e.g. \cite{fu2017dssd,lin2017focal, redmon2017yolo9000,redmon2018yolov3}. In this work, we adopt Faster R-CNN as the detection backbone for its robustness and flexibility.

\textbf{Domain Adaptation.}
As for the literature on domain adaptation, while it is quite vast, the most relevant category to our work is unsupervised domain adaptation in deep architectures. Recent works have involved discrepancy-based methods that guide the feature learning in DCNNs by minimizing the domain discrepancy with Maximum Mean Discrepancy (MMD) \cite{long2015learning,long2017deep,long2016unsupervised}. Another branch is to exploit the domain confusion by learning a domain discriminator \cite{ganin2014unsupervised,ganin2016domain,sankaranarayanan2017generate,tzeng2017adversarial}. Later, self-ensembling \cite{french2018self} extends Mean Teacher \cite{tarvainen2017mean} for domain adaptation and establishes new records on several cross-domain recognition benchmarks. All of the aforementioned works focus on the domain adaptation for recognition, and recently much attention has been paid to domain adaptation in other tasks, e.g., object detection \cite{chen2018domain,raj2015subspace} and semantic segmentation \cite{chen2018road,hoffman2016fcns,zhang2018fully}. For domain adaptation on object detection, \cite{siddiquie2013domain} uses transfer component analysis to learn the common transfer components across domains and \cite{raj2015subspace} aligns the region features with subspace alignment. More Recently, \cite{chen2018domain} constructs a domain adaptive Faster R-CNN by learning domain classifiers on both image and instance levels.

\textbf{Summary.} Similar to previous work \cite{chen2018domain}, our approach aims to leverage additional unlabeled target data for learning domain-invariant detector for cross-domain detection. The novelty is on the exploitation of Mean Teacher to bridge domain gap with consistency regularization in the context of object detection, which has not been previously explored. Moreover, the object relation between image regions is elegantly integrated into Mean Teacher paradigm to boost cross-domain detection.

\section{Mean Teacher in Semi-Supervised Learning}
We briefly review semi-supervised learning with Mean Teacher \cite{tarvainen2017mean}. Mean Teacher consists of two models with the same network architecture: a student model $f_\mathcal{S}$ parameterized by $w_{f_\mathcal{S}}$ and a teacher model $f_\mathcal{T}$ parameterized by $w_{f_\mathcal{T}}$. The main idea behind Mean Teacher is to encourage predictions of teacher and student consistent under small perturbations of inputs or network parameters. In other words, with the inputs of two different augmentations for the same unlabeled sample, teacher and student models should produce similar predicted probabilities. Specifically, in the standard setting of semi-supervised learning, we have access to labeled set $\mathcal{X}_L = \{(x_{l}, y_{l})\} $  and unlabeled set $\mathcal{X}_U = \{ x_{u}\} $. Given two perturbed samples $x_u^{\mathcal{S}}$ and $x_u^{\mathcal{T}}$ of the same unlabeled sample $x_{u}$, the consistency loss penalizes the difference between the student's prediction $f_\mathcal{S} (x_u^{\mathcal{S}};w_{f_\mathcal{S}}) $ and the teacher's $f_\mathcal{T} (x_u^{\mathcal{T}};w_{f_\mathcal{T}}) $, which is typically computed as the Mean Squared Error:
\begin{equation}\label{eq:Preliminaries:Mean:eq-2}
	\mathcal{L}_{cons} ( x_u ) = || f_\mathcal{S} (x_u^{\mathcal{S}};w_{f_\mathcal{S}}) - f_\mathcal{T} (x_u^{\mathcal{T}};w_{f_\mathcal{T}}) ||^2_{2}.
\end{equation}
The student is trained using gradient descent, while the weights of the teacher $w_{f_\mathcal{T}}$ at $t$-th iteration are the exponential moving average of the student weights $w_{f_\mathcal{S}}$: $w_{f_\mathcal{T}}^{t} = \alpha \cdot w_{f_\mathcal{T}}^{t-1} + (1 - \alpha)\cdot w_{f_\mathcal{S}}^{t-1} $. $\alpha$ is a smoothing coefficient parameter that controls the updating of teacher weights.

Hence, the total training loss in Mean Teacher is composed of supervised cross entropy loss on labeled samples and consistency loss of unlabeled samples, balanced with the tradeoff parameter $\lambda$:
\begin{equation}\label{eq:Preliminaries:Mean:eq-3}
	\mathcal{L} =\sum_{ (x_{l}, y_{l}) \in \mathcal{X}_L} L_{CE} ( x_{l}, y_{l} ) + \lambda \cdot \sum_{x_u \in \mathcal{X}_U } L_{cons} (x_u).
\end{equation}

\section{Mean Teacher in Cross-Domain Detection}
In this paper we remold Mean Teacher in the detection backbone (e.g., Faster R-CNN) for cross-domain detection by integrating the object relations into the measure of consistency regularization between teacher and student. An overview of our Mean Teacher with Object Relations (MTOR) framework is depicted in Figure \ref{fig:Framework}. We begin this section by elaborating the problem formulation. Then, a region-level consistency, which is  different from the generic consistency at image-level in primal Mean Teacher, is provided to facilitate domain adaptation at region-level. In addition, two kinds of graph-structured consistencies (inter-graph and intra-graph consistencies) are introduced to explore object relation in Mean Teacher, enabling the interaction between regions, which further enhance domain adaptation. Finally, the overall objective combining various consistencies along with its optimization strategy are provided.

\subsection{Problem Formulation}
In unsupervised domain adaptation, we are given $N_s$ labeled images $\mathcal{D}_s = \{(x_s,B_s)\}$ in source domain and $N_t$ unlabeled images $\mathcal{D}_t = \{x_t\}$ in target domain, where $B_s$ denotes the bounding box annotation for source image $x_s$. The ultimate goal of cross-domain detection is to design domain-invariant detectors depending on $\mathcal{D}_s$ and $\mathcal{D}_t$.

Inspired by the recent success of consistency-based methods in semi-supervised learning \cite{athiwaratkun2018improving,laine2016temporal, tarvainen2017mean} and Mean Teacher in cross-domain recognition \cite{french2018self}, we formulate our cross-domain detection model in a Mean Teacher paradigm by enforcing the predictions of teacher and student models consistent under perturbations of input unlabeled target sample. Accordingly, each labeled source sample $x_s$ is passed through student module to perform supervised learning of detection. Meanwhile, each unlabeled target sample $x_t$ is firstly transformed into two perturbed samples (i.e., $x_t^{\mathcal{T}}$ and $x_t^{\mathcal{S}}$) with different augmentations, and then fed into teacher and student models separately. This enables the measure of consistency between student and teacher. During training, different from Mean Teacher in cross-domain recognition \cite{french2018self} that solely encourages generic image-level consistency, we consider the consistency at a finer granularity (i.e., region-level), which is tailored for object detection. Moreover, two graph-structured consistencies are especially designed to exploit object relations in the context of Mean Teacher, which further boosts adaptation by aligning the results depending on the inherent relations between objects.

Specifically, given the identical set of region proposals $\mathcal{R}_{x_t}^{\mathcal{T}}=\{r_t\}$ generated by teacher model $F^{\mathcal{T}}$, we construct two relational graphs $\mathcal{G}^{\mathcal{T}}_{x_t}$ and $\mathcal{G}^{\mathcal{S}}_{x_t}$ to learn the affinity matrix that captures the relation between any pair of regions in teacher and student, respectively. Note that we use $\mathcal{G}_{x_t} \in \{\mathcal{G}^{\mathcal{T}}_{x_t},\mathcal{G}^{\mathcal{S}}_{x_t}\}$ for simplicity, i.e., $\mathcal{G}_{x_t}$ denotes the graph in either teacher $\mathcal{G}^{\mathcal{T}}_{x_t}$ or student $\mathcal{G}^{\mathcal{S}}_{x_t}$. More precisely, by treating each region in teacher/student as one vertex, the relational graph is constructed as $\mathcal{G}_{x_t} = \{\mathcal{V}_{x_t}, \mathcal{E}_{x_t}\}$ , where $\mathcal{V}_{x_t}$ denotes the set of predictions for all region proposals in teacher/student and $\mathcal{E}_{x_t}$ is a ($|\mathcal{V}_{x_t}| \times |\mathcal{V}_{x_t}|$) affinity matrix whose entry measures the similarities between every two regions. $\mathcal{E}_{x_t}$ is symmetric, and represents an undirected weighted graph. On the basis of two constructed relational graphs, we make the detection backbone---Faster R-CNN transferable across domains in Mean Teacher paradigm with three consistency regularization: 1) region-level consistency (Section \ref{sec:Cross:Region}) to align the region-level predictions of the vertices in teacher and student graphs sharing the same spatial location, 2) inter-graph consistency (Section \ref{sec:Cross:Inter-Graph}) for matching the graph structures (i.e., the affinity matrices) of teacher and student graphs, and 3) intra-graph consistency (Section \ref{sec:Cross:Intra-Graph}) to enhance the similarity between regions belonging to the same class within the graph of student.

\subsection{Region-Level Consistency} \label{sec:Cross:Region}
Unlike  \cite{french2018self} that pursues image-level consistency to perturbations of inputs in recognition, we facilitate Mean Teacher in cross-domain detection by exploiting region-level consistency under the identical region proposals between teacher and student. The design of region-level consistency helps to reduce the local instance variances such as scale, color jitter, random noise, \emph{etc}, which in turn implicitly enforces the consistency of object localization.

Technically, given the two perturbed samples $x_t^{\mathcal{T}}$ and $x_t^{\mathcal{S}}$ of one unlabeled target sample $x_t$, they are fed into teacher and student detectors under the same backbone (i.e., Faster R-CNN) separately. Faster R-CNN is a two-stage detector consisting of three major components: a Base Convolution Neural Network (Base CNN) for feature extraction, a Region Proposal Network (RPN) to generate candidate region proposals, and a Region-based Convolution Neural Network (RCNN) for classifying each region. Hence, with the input of $x_t^{\mathcal{T}}$, the Base CNN of teacher $F_{Conv}^{\mathcal{T}}$ firstly produces output feature map  $\textit{\textbf{f}}_{x_t}^{\mathcal{T}}$. Next, depending on the output feature map $\textit{\textbf{f}}_{x_t}^{\mathcal{T}}$, a set of region proposals $\mathcal{R}_{x_t}^{\mathcal{T}}=\{r_t\}$ are generated via RPN in teacher $F_{RPN}^{\mathcal{T}}$:
\vspace{-0.02in}
\begin{equation}\label{eq:Cross:Region:eq-1}\small
	\textit{\textbf{f}}_{x_t}^{\mathcal{T}}  = F_{Conv}^{\mathcal{T}} (x_t^{\mathcal{T}}), \mathcal{R}_{x_t}^{\mathcal{T}} = F_{RPN}^{\mathcal{T}} (\textit{\textbf{f}}_{x_t}^{\mathcal{T}}).
	\vspace{-0.02in}
\end{equation}
For each region proposal $r_t \in \mathcal{R}_{x_t}^{\mathcal{T}}$, a ROI pooling layer~is utilized to extract a fixed-length vector $\textit{\textbf{f}}^{\mathcal{T}}_{r}$ from the feature map $\textit{\textbf{f}}_{x_t}^{\mathcal{T}}$, which represents the region feature of $r_t$ in teacher. The RCNN in teacher $ F_{RCNN}^{\mathcal{T}}$ further takes each region feature $\textit{\textbf{f}}^{\mathcal{T}}_{r}$ as input and classifies it into one of the $C$ foreground categories and a catch-all background class. Here the prediction of each region is the probability distribution over background plus foreground categories, which is denoted as $\textit{\textbf{d}}^{\mathcal{T}}_{r} =  F_{RCNN}^{\mathcal{T}} ( \textit{\textbf{f}}^{\mathcal{T}}_{r}) $.
As such, by accumulating the predicted results of all region proposals, the entire detection output of $x_t^{\mathcal{T}}$ in teacher is denoted as $\mathcal{V}_{x_t}^{\mathcal{T}} = \{ \textit{\textbf{d}}^{\mathcal{T}}_{r}\}$. Similarly, for student model $F^{\mathcal{S}}$, another perturbed image $x_t^{\mathcal{S}}$ is fed into its Base CNN $F_{Conv}^{\mathcal{S}}$ to produce the feature map $\textit{\textbf{f}}_{x_t}^{\mathcal{S}}$. Note that instead of generating another set of region proposals for $x_t^{\mathcal{S}}$ via RPN in student, we directly take the region proposals from teacher $\mathcal{R}_{x_t}^{\mathcal{T}}$ as the ones in~student:
\begin{equation}\label{eq:Cross:Region:eq-2}\small
	\textit{\textbf{f}}_{x_t}^{ \mathcal{S}}  = F_{Conv}^{\mathcal{S}} (x_t^{ \mathcal{S}}), \mathcal{R}_{x_t}^{ \mathcal{S}} = \mathcal{R}_{x_t}^{\mathcal{T}}.
\end{equation}
That is, we endow teacher and student with the same set of region proposals, enabling the interaction between teacher and student for measuring region-level consistency. Given region proposals $\mathcal{R}_{x_t}^{ \mathcal{S}}$ and feature map $\textit{\textbf{f}}_{x_t}^{ \mathcal{S}}$, we can acquire the region feature $\textit{\textbf{f}}^{\mathcal{S}}_{r}$ for each region proposal and the corresponding probability distribution $\textit{\textbf{d}}^{\mathcal{S}}_{r} =  F_{RCNN}^{\mathcal{S}} ( \textit{\textbf{f}}^{\mathcal{S}}_{r})$, leading to the entire detection results in student $\mathcal{V}_{x_t}^{\mathcal{S}} = \{ \textit{\textbf{d}}^{\mathcal{S}}_{r}\}$.

As such, the region-level consistency is measured as the distance between the prediction of teacher $\mathcal{V}_{x_t}^{\mathcal{T}}$ and that of student $\mathcal{V}_{x_t}^{\mathcal{S}}$. To focus more on foreground samples and stabilize the training in the challenging cross-domain detection scenario, we follow \cite{french2018self} and adopt confidence thresholding to filter out background region proposals and low-confidence foreground region proposals with noise. For each region proposal $r_t\in\mathcal{R}_{x_t}^{\mathcal{T}}$ of teacher model, we compute the confidence as $q^{\mathcal{T}}_{r} =  \max\limits\nolimits_{j \in \mathcal{C}}(\textit{\textbf{d}}^{\mathcal{T}}_{rj})$, where $\mathcal{C}$ is the set of $C$ foreground categories and $\textit{\textbf{d}}^{\mathcal{T}}_{rj}$ is the predicted probability of $j$-th foreground category. If $q^{\mathcal{T}}_{r}$ is below the confidence threshold $\epsilon$, we eliminate the region proposal in $\mathcal{R}_{x_t}^{\mathcal{T}}$. With the refined region proposal ($\mathcal{R}_{x_t}^{\mathcal{T}}$), and the corresponding region-level predictions of teacher and student ($\mathcal{V}_{x_t}^{\mathcal{T}} = \{ \textit{\textbf{d}}^{\mathcal{T}}_{r}\}$ and $\mathcal{V}_{x_t}^{\mathcal{S}} = \{ \textit{\textbf{d}}^{\mathcal{S}}_{r}\}$), the \textbf{R}egion-level \textbf{C}onsistency \textbf{L}oss (RCL) is calculated as the average of Mean Squared Error between the region-level predictions of teacher and student for all region proposals:
\begin{equation}\label{eq:Cross:Region:eq-5}\small
	L^{RCL}_{x_t} = \frac{ 1 }{|R_{x_t}^{\mathcal{T}}|} \cdot \sum_{r \in R_{x_t}^{\mathcal{T}}} ||\textit{\textbf{d}}^{\mathcal{T}}_{r}  - \textit{\textbf{d}}^{\mathcal{S}}_{r} ||^2_{2}.
\end{equation}

\subsection{Inter-Graph Consistency} \label{sec:Cross:Inter-Graph}
The region-level consistency only individually aligns the predictions of each region proposal in teacher and student, while leaving the relations between regions unexploited. Thus, inspired from graph structure exploitation \cite{pan2016learning,pan2014click,yao2018exploring,yao2015semi} in computer vision tasks, we devise a novel graph-structured regularization, i.e., inter-graph consistency, to measure the consistency of graph structures under perturbations of inputs by matching the affinity matrices of graphs constructed in teacher and student models. The rationale of inter-graph consistency is that the inherent relations between objects within each image should be invariant to different image augmentations.

In particular, for the graph constructed in teacher $\mathcal{G}^{\mathcal{T}}_{x_t} = \{\mathcal{V}^{\mathcal{T}}_{x_t}, \mathcal{E}^{\mathcal{T}}_{x_t}\}$, the affinity matrix of teacher $\mathcal{E}^{\mathcal{T}}_{x_t}$ is obtained by defining each entry as the similarity between two regions. For instance, given two region proposals $r_m,r_n \in \mathcal{R}_{x_t}^{\mathcal{T}}$, the entry $(\mathcal{E}_{x_t}^{\mathcal{T}})_{{m,n}}$ in $\mathcal{E}^{\mathcal{T}}_{x_t}$ is calculated as the cosine similarity between the region representations ($\textit{\textbf{f}}^{\mathcal{T}}_{r_m}$ and $\textit{\textbf{f}}^{\mathcal{T}}_{r_n}$):
\begin{equation}\label{eq:Cross:Inter-Graph:eq-1}\small
	(\mathcal{E}_{x_t}^{\mathcal{T}})_{{m,n}}  = \frac{\textit{\textbf{f}}^{\mathcal{T}}_{r_m} \cdot \textit{\textbf{f}}^{\mathcal{T}}_{r_n} }{ || \textit{\textbf{f}}^{\mathcal{T}}_{r_m} ||_{2} \cdot||\textit{\textbf{f}}^{\mathcal{T}}_{r_n} ||_{2}}.
\end{equation}
Similarly, we achieve the affinity matrix of student $\mathcal{E}_{ x_t^{\mathcal{S}} }$ by measuring the cosine similarities between every two regions in student. Accordingly, the Int\textbf{E}r-\textbf{G}raph Consistency \textbf{L}oss (EGL) is defined as the Mean Squared Error between the affinity matrices of graphs in teacher and student models:
\begin{equation}\label{eq:Cross:Inter-Graph:eq-2}\small
	L^{EGL}_{x_t} = \frac{ 1 }{|R_{x_t}^{\mathcal{T}}|^2} \cdot||\mathcal{E}^{\mathcal{S}}_{ x_t} - \mathcal{E}^{\mathcal{T}}_{ x_t}||^2_{2}.
\end{equation}

\subsection{Intra-Graph Consistency in Student} \label{sec:Cross:Intra-Graph}
Inspired from self-labeling \cite{lee2013pseudo,saito2017asymmetric} for domain adaptation, the inter-graph consistency is devised to further reinforce the similarity between regions of same class within the graph of student with the supervision from teacher. Specifically, since no label is provided for target samples in unsupervised domain adaptation settings, we directly utilize the teacher to assign each region proposal $r_t \in \mathcal{R}_{x_t}^{\mathcal{T}}$ a ``pseudo" label: $\hat{l}_{r} =  arg\max\limits\nolimits_{j \in C} (\textit{\textbf{d}}^{\mathcal{T}}_{rj})$. Next, a ($|\mathcal{R}_{x_t}^{\mathcal{T}}| \times |\mathcal{R}_{x_t}^{\mathcal{T}}|$) supervision matrix $M^{\mathcal{T}}_{x_t}$ is naturally generated to indicate whether two regions belong to the same category:
\begin{equation}\label{eq:Cross:Intra-Graph:eq-5}\small
	(M^{\mathcal{T}}_{x_t})_{(m,n)} =
	\begin{cases}
		1 & \text{if }\hat{l}_{r_m} = \hat{l}_{r_n}, \\
		0 & \text{otherwise}.
	\end{cases},
\end{equation}
where $\hat{l}_{r_m}$ and $\hat{l}_{r_n}$ denote the pseudo labels of two regions $r_m,r_n \in \mathcal{R}_{x_t}^{\mathcal{T}}$, respectively. Thus, given the the affinity matrix of student $\mathcal{E}^{\mathcal{S}}_{ x_t }$ and the supervision matrix $M^{\mathcal{T}}_{x_t}$, the intr\textbf{A}-\textbf{G}raph consistency \textbf{L}oss (AGL) is defined as:
\begin{equation}\label{eq:Cross:Intra-Graph:eq-6}\small
	L^{AGL}_{x_t} =\frac{  \sum\limits_{{1 \le m,n \le |\mathcal{R}_{x_t}^{\mathcal{T}}|}} (M^{\mathcal{T}}_{x_t})_{(m,n)} \cdot (1 - (\mathcal{E}^{\mathcal{S}}_{ x_t })_{(m,n)} ) }{  max(1,\sum\limits_{1 \le m,n \le |\mathcal{R}_{x_t}^{\mathcal{T}}|} (M^{\mathcal{T}}_{x_t})_{(m,n)})}.
\end{equation}
Note that $L^{AGL}_{x_t}$ is triggered when at least two regions share the same pseudo label in $\mathcal{R}_{x_t}^{\tau}$. By minimizing the inter-graph consistency loss, the similarity between regions with the same pseudo label in student is enhanced, pursuing lower intra-class variation within the graph of student.

\subsection{Optimization}
\textbf{Training Objective.}
The overall training objective of our MTOR integrates the supervised loss $\mathcal{L}_{sup}$ on labeled source data $\mathcal{D}_{s}$ and three consistency losses, i.e., region-level consistency $L^{RCL}_{x_t}$ in Eq.(\ref{eq:Cross:Region:eq-5}), inter-graph consistency $L^{EGL}_{x_t}$ in Eq.(\ref{eq:Cross:Inter-Graph:eq-2}) and intra-graph consistency $L^{AGL}_{x_t}$ in Eq.(\ref{eq:Cross:Intra-Graph:eq-6}) on unlabeled target data $\mathcal{D}_{t}$:
\begin{equation}\label{eq:OPT:Loss:eq-1}\small
	\mathcal{L}  = \sum_{ (x_s, B_s) \in \mathcal{D}_{s} } \mathcal{L}_{sup}( x_s, B_s) + \lambda \cdot \sum_{ x_t \in \mathcal{D}_{t} } ( L^{RCL}_{x_t} + L^{EGL}_{x_t} + L^{AGL}_{x_t} ),
\end{equation}
where $\lambda$ is the tradeoff parameter.

\textbf{Weights Update.}
The student network $F^{\mathcal{S}}$ is optimized with standard SGD algorithm by minimizing $\mathcal{L}$. The weights of teacher network $F^{\mathcal{T}}$ at iteration $t$ are updated as the exponential moving average of student weights:
\begin{equation}\label{eq:OPT:Loss:eq-2}\small
	{w}_{F^{\mathcal{T}}}^{t}  = \alpha \cdot {w}_{F^{\mathcal{T}}}^{t-1} + (1 - \alpha) \cdot {w}_{F^{\mathcal{S}}}^{t-1},
\end{equation}
where $\alpha$ denotes smoothing coefficient parameter.

\section{Experiments}
We conduct extensive evaluations of our MTOR for cross-domain detection in two different domain shift scenarios, including one normal-to-foggy weather transfer in urban scene (\textit{Cityscapes} \cite{cordts2016cityscapes} $\rightarrow$ \textit{Foggy Cityscapes} \cite{sakaridis2018semantic}) and two synthetic-to-real transfers (i.e., \textit{SIM10k} \cite{johnson2016driving} $\rightarrow$ \textit{Cityscapes} and 3D CAD-rendered images $\rightarrow$ real images in \textit{Syn2Real} detection dataset \cite{peng2018syn2real}).

\subsection{Dataset and Experimental Settings}
\textbf{Dataset.}
The \textit{Cityscapes} dataset (\textbf{C}) is a popular semantic understanding benchmark in urban street scenes with pixel-level annotation, containing 2,975 images for training and 500 images for validation. Since it is not dedicated for detection, we follow \cite{chen2018domain} and generate the bounding box annotations by the tightest rectangles of each instance segmentation mask for 8 categories (\textit{person} plus 7 kinds of \textit{transports}). \textit{Foggy Cityscapes} (\textbf{F}) is a recently proposed synthetic foggy dataset which simulates fog on real scenes. Each foggy image is rendered with clear image and depth map from \textit{Cityscapes}. Thus the annotations and data split in \textit{Foggy Cityscapes} are inherited from \textit{Cityscapes}. \textit{SIM10k} (\textbf{M}) dataset contains 10$k$ images rendered from computer game---Grand Theft Auto V (GTA 5) with bounding box annotations for \textit{cars}. The \textit{Syn2Real} detection dataset is the largest synthetic-to-real object detection dataset to date with over 70$k$ images in the training, validation and testing domains. The training domain consists of 8$k$ synthetic images (\textbf{S}) which are generated from 3D CAD models. Each object is rendered independently and placed on a white background. The validation domain includes 3,289 real images from \textit{COCO} \cite{lin2014microsoft} (\textbf{O}) and the testing domain contains 60,863 images from video frames in \textit{YTBB} \cite{real2017youtube} (\textbf{Y}).

\textbf{Normal-to-Foggy Weather Transfer.}
We follow \cite{chen2018domain} and evaluate \textbf{C} $\rightarrow$ \textbf{F} for transfer across different weather conditions. The training set in \textit{Cityscapes} is taken as source domain. We use the training set in \textit{Foggy Cityscapes} as target domain and results are reported on its validation set.

\textbf{Synthetic-to-Real Image Transfer.}
We consider two directions for synthetic-to-real transfers: \textbf{M} $\rightarrow$ \textbf{C} and \textbf{S} $\rightarrow$ \textbf{O}/\textbf{Y}. For \textbf{M} $\rightarrow$ \textbf{C}, we utilize the entire \textit{SIM10k} as source domain and leverage \textit{Cityscapes} training set as target domain. The results are reported on \textit{Cityscapes} validation split. For \textbf{S} $\rightarrow$ \textbf{O}/\textbf{Y} on \textit{Syn2Real} detection dataset, we take the training set (synthetic images) as source domain and the validation set (\textit{COCO})/testing set (\textit{YTBB}) as target domain. Since the annotations of testing set are not publicly available, we submit results to online testing server for evaluation.

\textbf{Implementation Details.}
For \textbf{C} $\rightarrow$ \textbf{F} and \textbf{M} $\rightarrow$ \textbf{C}, we adopt the 50-layer ResNet \cite{he2016resnet} pre-trained on ImageNet \cite{ILSVRC15} as the basic architecture of Faster R-CNN backbone. For the more challenging \textbf{S} $\rightarrow$ \textbf{O}/\textbf{Y}, the Faster R-CNN backbone is mainly constructed on 152-layer ResNet. For all transfers, we utilize ``image-centric" sampling strategy \cite{girshick2015fast}. Each input image is resized such that its scale (shorter edge) is 600 pixels. Each mini-batch contains 2 images per GPU, one from the source domain and the other from the target domain. We train on 4 GPUs (so effective mini-batch size is 8) and each image has 128 sampled anchors, with a ratio of 1:3 of positive to negatives \cite{girshick2015fast}. We implement MTOR based on MXNet \cite{chen2015mxnet}. Specifically, the network weights are trained by SGD optimizer with 0.0005 weight decay and 0.9 momentum. The learning rate and maximum training epoch are set as 0.001 and 10 for all experiments. The confidence threshold $\epsilon$ is empirically set to 0.98 for \textbf{C} $\rightarrow$ \textbf{F} and \textbf{M} $\rightarrow$ \textbf{C}, and 0.99 for \textbf{S} $\rightarrow$ \textbf{O}/\textbf{Y}. The tradeoff parameter $\lambda$ in Eq.(\ref{eq:OPT:Loss:eq-1}) and the smooth coefficient parameter $\alpha$ in Eq.(\ref{eq:OPT:Loss:eq-2}) is set as 1.0 and 0.99, respectively. Moreover, our MTOR is firstly pre-trained on labeled source data. For data augmentations on target images, we firstly augment each target image with the same spatial perturbation including random cropping, padding, or flipping. Next, we additionally perform two different kinds of image augmentations with random color jittering (i.e., brightness, contrast, hue and saturation augmentations) or PCA noise, resulting in two perturbed target samples, one for student and the other for teacher. Following \cite{chen2018domain}, we report mAP with a IoU threshold of 0.5 for evaluation.

\textbf{Compared Approaches.}
To empirically verify the merit of our MTOR, we compare the following methods: (1) \textbf{Source-Only} directly exploits the Faster R-CNN model trained on source domain to detect objects in target samples. (2) \textbf{DA}\cite{chen2018domain} designs two domain classifiers to alleviate both image-level and region-level domain discrepancy, which are further enforced with a consistency regularizer. (3) \textbf{MTOR} is the proposal in this paper. Moreover, we design three degraded variants trained with region-level consistency ($\textbf{MTOR}_{R}$), region-level plus inter-graph consistency ($\textbf{MTOR}_{RE}$), and region-level plus intra-Graph consistency ($\textbf{MTOR}_{RA}$). (4) \textbf{Train-on-target} is an oracle run that trains Faster R-CNN on all the labeled target samples.

\begin{table*}\small
	\center
	\vspace{-0.00in}
	\setlength{\tabcolsep}{8pt}
	\caption{The mean Average Precision (mAP) of different models on \textit{Foggy Cityscapes} validation set for \textbf{C} $\rightarrow$ \textbf{F} transfer.}
	\vspace{-0.10in}
	\begin{tabular}{ l | c c c | c c c c c c c c | c}
		\hlineB{2}
		                         & \textbf{RCL} & \textbf{EGL} & \textbf{AGL} & person        & rider         & car           & truck         & bus           & train         & mcycle        & bicycle       & mAP           \\ \hline \hline
		Source-only              &              &              &              & 25.7          & 35.9          & 36.0          & 19.4          & 30.8          & 9.7           & \textbf{29.0} & 28.9          & 26.9          \\ \hline
		DA \cite{chen2018domain} &              &              &              & 29.2          & 40.4          & 43.4          & 19.7          & 38.3          & 28.5          & 23.7          & 32.7          & 32.0          \\ \hlineB{2}
										
		$ \text{MTOR}_{R}$       & \checkmark   &              &              & \textbf{30.8} & \textbf{41.5} & 44.1          & 21.6          & 37.8          & 35.1          & 26.7          & \textbf{35.8} & 34.2          \\ \cline{2-13}
		$ \text{MTOR}_{RE}$      & \checkmark   & \checkmark   &              & 28.7          & 40.1          & \textbf{45.9} & \textbf{22.9} & 38.0          & 38.6          & 26.9          & 34.9          & 34.5          \\ \cline{2-13}
		$ \text{MTOR}_{RA}$      & \checkmark   &              & \checkmark   & 29.6          & 41.2          & 43.7          & 22.2          & 38.4          & \textbf{40.9} & 27.8          & 35.3          & 34.9          \\ \cline{2-13}
		$ \text{MTOR}$           & \checkmark   & \checkmark   & \checkmark   & 30.6          & 41.4          & 44.0          & 21.9          & \textbf{38.6} & 40.6          & 28.3          & 35.6          & \textbf{35.1} \\ \hlineB{2}
		Train-on-target          &              &              &              & 31.4          & 42.6          & 51.7          & 28.8          & 43.4          & 40.2          & 31.7          & 33.2          & 37.9          \\ \hlineB{2}
	\end{tabular}
	\vspace{-0.25in}
	\label{tab:eval_foggy}
\end{table*}

\subsection{Performance Comparison and Analysis}
\textbf{Normal-to-Foggy Weather Transfer.} Table \ref{tab:eval_foggy} shows the performance comparisons on \textit{Foggy Cityscapes} validation set for \textbf{C} $\rightarrow$ \textbf{F} transfer. Overall, the results with regard to mAP score indicate that our proposed MTOR achieves superior performance against state-of-the-art technique (DA). In particular, the mAP of MTOR can achieve 35.1\%, making 3.1\% absolute improvement over the best competitor DA. The performances of Source-only which trains Faster R-CNN only on the labeled source data can be regarded as a lower bound without adaptation. By additionally incorporating the domain classifier in both image and region level, DA leads to a large performance boost over Source-only, which basically indicates the advantage of alleviating the domain discrepancy over the source and target data. Note that for fair comparison, we re-implemented DA based on the same 50-layer ResNet architecture. However, the performances of DA are still lower than our $\text{MTOR}_{R}$, which utilizes region-level consistency regularization in Mean Teacher paradigm. This confirms the effectiveness of enforcing region-level consistency under perturbations of unlabeled target samples for cross-domain detection. In addition, by further integrating object relations into Mean Teacher paradigm through graph-structured consistency from inter-graph or intra-graph perspective, our $\text{MTOR}_{RE}$ and $ \text{MTOR}_{RA}$ improve $\text{MTOR}_{R}$. The results demonstrate the advantage of inter-graph consistency to match the graph structures between teacher and student, and intra-graph consistency to enhance the similarity between regions of same class in student. By simultaneously utilizing region-level and two graph-structured consistencies, MTOR further boosts up the performances, which indicates the merit of jointly exploiting inter-graph and intra-graph consistencies in Mean Teacher paradigm.

\begin{table}\small
	\center
	\caption{The Average Precision (AP) of \textit{car} on \textit{Cityscapes} validation set for \textbf{M} $\rightarrow$ \textbf{C} transfer.}
	\begin{tabular}{ l | c c c | c}
		\hlineB{2}
		                         & \textbf{RCL} & \textbf{EGL} & \textbf{AGL} & car AP        \\ \hline \hline
		Source-only              &              &              &              & 39.4          \\ \hline
		DA \cite{chen2018domain} &              &              &              & 41.9          \\ \hlineB{2}
		$ \text{MTOR}_{R}$       & \checkmark   &              &              & 45.9          \\ \cline{2-5}
		$ \text{MTOR}_{RE}$      & \checkmark   & \checkmark   &              & 46.1          \\ \cline{2-5}
		$ \text{MTOR}_{RA}$      & \checkmark   &              & \checkmark   & 46.3          \\ \cline{2-5}
		$ \text{MTOR}$           & \checkmark   & \checkmark   & \checkmark   & \textbf{46.6} \\ \hlineB{2}
		Train-on-target          &              &              &              & 58.6          \\ \hlineB{2}
	\end{tabular}
	\vspace{-0.3in}
	\label{tab:eval_sim10k}
\end{table}

\begin{table*}\small
	\center
	\vspace{-0.03in}
	\setlength{\tabcolsep}{3pt}
	\caption{The mean Average Precision (mAP) of different models on \textit{Syn2Real} detection dataset for \textbf{S} $\rightarrow$ \textbf{O}/\textbf{Y} transfers.}
	\begin{tabular}{ l | c c c | c c c c c c c c c c c c | c}
		\hlineB{2}
		                         & \textbf{RCL} & \textbf{EGL} & \textbf{AGL} & plane         & bcycl         & bus           & car           & horse         & knife         & mcycl         & person        & plant         & sktbd         & train         & truck        & mAP           \\ \hline \hline
		\multicolumn{16}{l}{\textit{mAP on validation set (COCO) for \textbf{S} $\rightarrow$ \textbf{O} transfer:} } \\ \hline
		Source-only              &              &              &              & 30.0          & \textbf{25.3} & 31.3          & 14.0          & 17.3          & 1.9           & 25.6          & 18.5          & 14.7          & 14.7          & 21.1          & 2.2          & 18.1          \\ \hline
		DA \cite{chen2018domain} &              &              &              & 30.3          & 24.1          & 31.3          & 14.0          & 17.4          & 1.3           & 27.4          & 18.9          & \textbf{17.5} & 14.5          & 21.8          & \textbf{3.1} & 18.5          \\ \hlineB{2}
		$ \text{MTOR}_{R}$       & \checkmark   &              &              & 32.0          & 22.8          & 29.1          & 15.3          & \textbf{20.8} & 0.6           & \textbf{32.4} & 22.2          & 0.5           & 18.2          & \textbf{36.9} & 0.6          & 19.3          \\ \cline{2-17}
		$ \text{MTOR}_{RE}$      & \checkmark   & \checkmark   &              & 33.3          & 21.2          & 32.9          & 13.1          & 18.1          & \textbf{3.1}  & 32.2          & 24.0          & 1.4           & \textbf{20.5} & 34.4          & 0.6          & 19.6          \\ \cline{2-17}
		$ \text{MTOR}_{RA}$      & \checkmark   &              & \checkmark   & 35.4          & 24.0          & 32.1          & 14.9          & 19.1          & 1.8           & 31.6          & \textbf{24.2} & 3.7           & 18.9          & 31.7          & 2.0          & 20.0          \\ \cline{2-17}
		$ \text{MTOR}$           & \checkmark   & \checkmark   & \checkmark   & \textbf{35.5} & 24.9          & \textbf{32.9} & \textbf{15.4} & 19.1          & 1.8           & 31.4          & 21.8          & 14.4          & 18.9          & 30.4          & 1.7          & \textbf{20.7} \\ \hlineB{2}
		Train-on-target          &              &              &              & 84.5          & 52.2          & 77.5          & 58.7          & 76.1          & 28.9          & 65.4          & 71.9          & 49.2          & 70.5          & 83.8          & 52.5         & 64.3          \\ \hline \hline
		\multicolumn{16}{l}{\textit{mAP on official testing set (YTBB) for \textbf{S} $\rightarrow$ \textbf{Y} transfer:} } \\ \hline
		Source-only              &              &              &              & 28.4          & 18.4          & 23.8          & 28.4          & 35.8          & 3.6           & 35.7          & \textbf{8.6}           & 8.4           & 14.8          & 6.4           & 5.2          & 18.1          \\ \hline
		DA \cite{chen2018domain} &              &              &              & 38.0          & 16.1          & 23.3          & 30.7          & 33.0          & 4.7           & 34.8          & 6.1           & \textbf{15.7}          & 14.0          & \textbf{9.8}           & \textbf{9.5}          & 19.6          \\ \hlineB{2}
		MTOR (Ours)              & \checkmark   & \checkmark   & \checkmark   & \textbf{42.8} & \textbf{21.0} & \textbf{31.3} & \textbf{33.3} & \textbf{42.9} & \textbf{10.2} & \textbf{38.5} & 7.2  & 12.9 & \textbf{18.0} & 7.2  & 8.2 & \textbf{22.8} \\ \hlineB{2}
	\end{tabular}
	\vspace{-0.25in}
	\label{tab:eval_visda}
\end{table*}

\textbf{Synthetic-to-Real Image Transfer.}
The performance comparisons for synthetic-to-real transfer task on \textbf{M} $\rightarrow$ \textbf{C} are summarized in Table \ref{tab:eval_sim10k}. Our MTOR exhibits better performance than other runs. In particular, the AP of \textit{car} for MTOR can reach $46.6\%$, making the absolute improvement over DA by $4.7\%$. Similar to the observations in normal-to-foggy weather transfer, $\text{MTOR}_{R}$ performs better than DA by aligning region-level predictions in Mean Teacher and the performance is further improved by incorporating inter-graph and intra-graph consistency in $\text{MTOR}_{RE}$ and $\text{MTOR}_{RA}$. Combining all the three consistency regularizations, our MTOR achieves the best performance.

\begin{figure}[!tb]
	\centering {\includegraphics[width=0.43\textwidth]{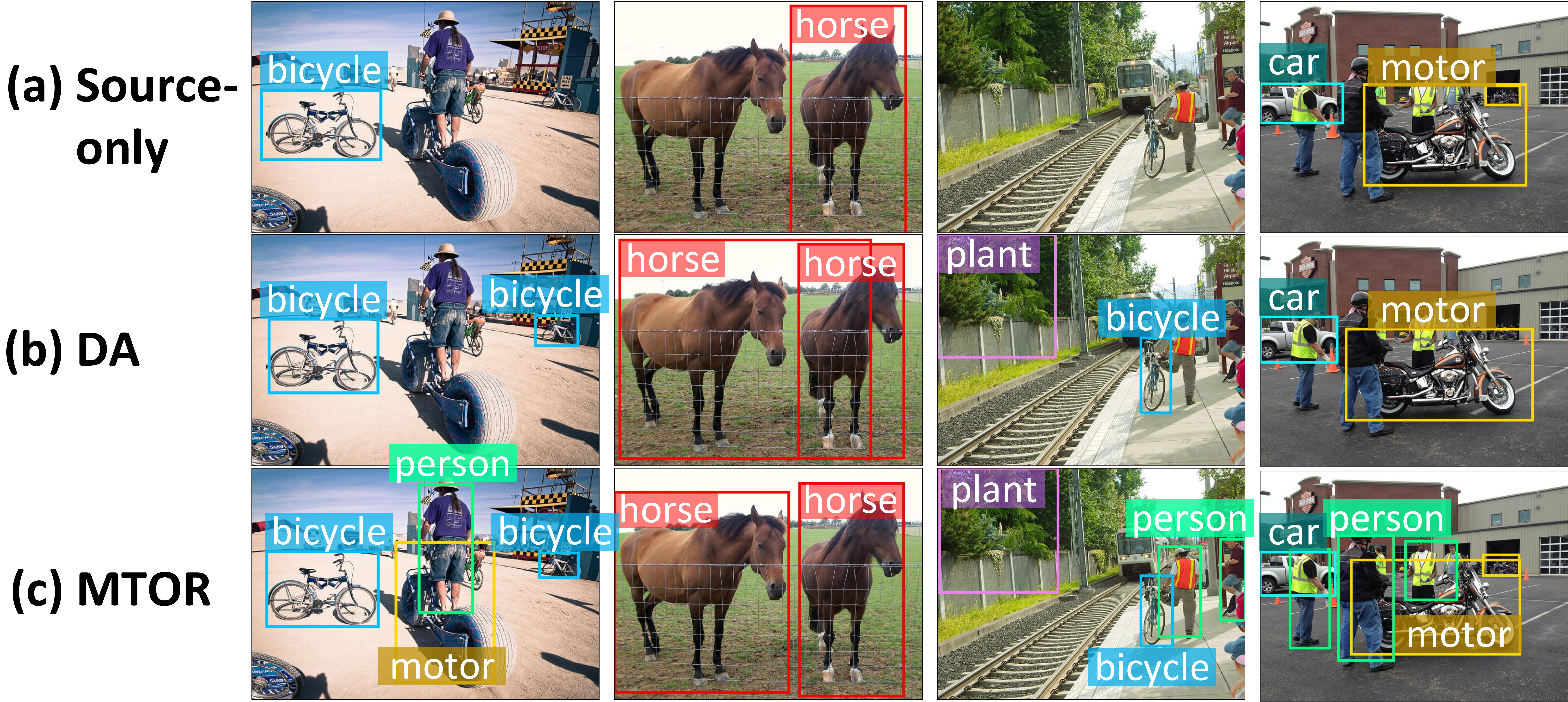}}
	\caption{\small Examples of detection results on \textit{COCO} for \textbf{S} $\rightarrow$ \textbf{O}.}
	\vspace{-0.3in}
	\label{fig:vis}
\end{figure}

We further evaluate our approach for \textbf{S} $\rightarrow$ \textbf{O}/\textbf{Y} transfer on the more challenging \textit{Syn2Real} detection dataset. Table \ref{tab:eval_visda} shows the performance comparisons on \textbf{S} $\rightarrow$ \textbf{O} transfer. A clear performance improvement is achieved by our proposed MTOR over other baselines. Similar to the observations on the transfers across \textit{SIM10k}, \textit{Cityscapes}, and \textit{Foggy Cityscapes}, $\text{MTOR}_{R}$ performs better than DA by taking region-level consistency on target samples into account for cross-domain detection. Moreover, $\text{MTOR}_{RE}$ and $\text{MTOR}_{RA}$ exhibit better performance than $\text{MTOR}_{R}$ by additionally pursuing inter-graph and intra-graph consistency respectively, and further performance improvement is attained when exploiting region-level consistency plus two graph-structured consistencies by MTOR. We also submitted our MTOR, Source-only, and DA to online evaluation server and evaluated the performances on official testing set. Table \ref{tab:eval_visda} summaries the performances on official testing set \textit{YTBB} for \textbf{S} $\rightarrow$ \textbf{Y} transfer. The results clearly show that our MTOR outperforms two other baselines.

\textbf{Qualitative Analysis.} Figure \ref{fig:vis} showcases four examples of detection results on \textit{COCO} for \textbf{S} $\rightarrow$ \textbf{O} transfer by three approaches, i.e., Source-only, DA and our MTOR. The exemplar results clearly show that our MTOR can generate more accurate detection results by exploring region-level and graph-structured consistency in Mean Teacher paradigm to boost cross-domain detection. For instance, MTOR correctly detects \textit{person} in the fourth image which is missed in Source-only and DA.

\begin{figure}
	\centering
	\vspace{-0.05in}
	\subfigure{
		\label{fig:exp:loss_lambda}
		\includegraphics[width=0.2\textwidth]{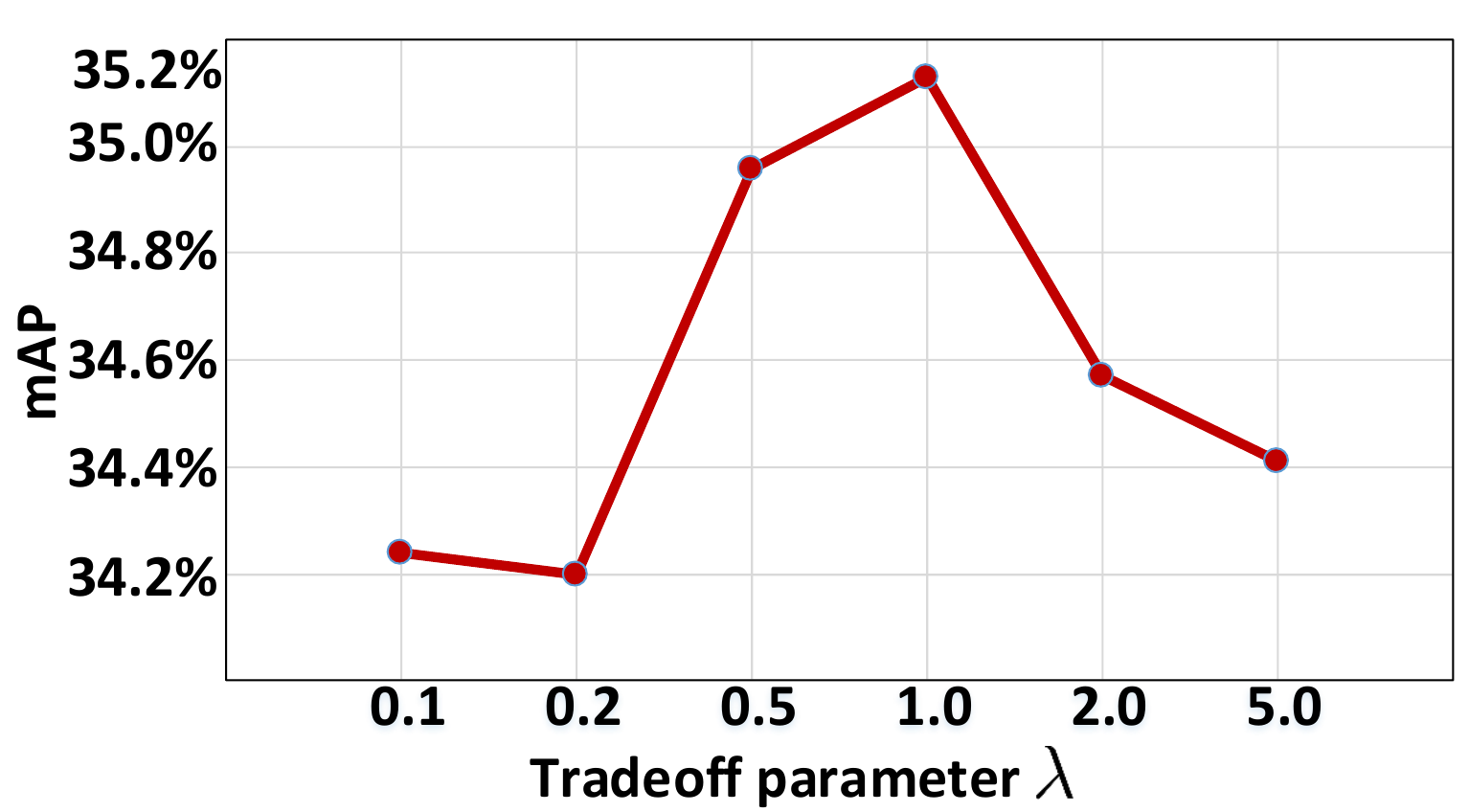}}
	\subfigure{
		\label{fig:exp:ema_alpha}
		\includegraphics[width=0.2\textwidth]{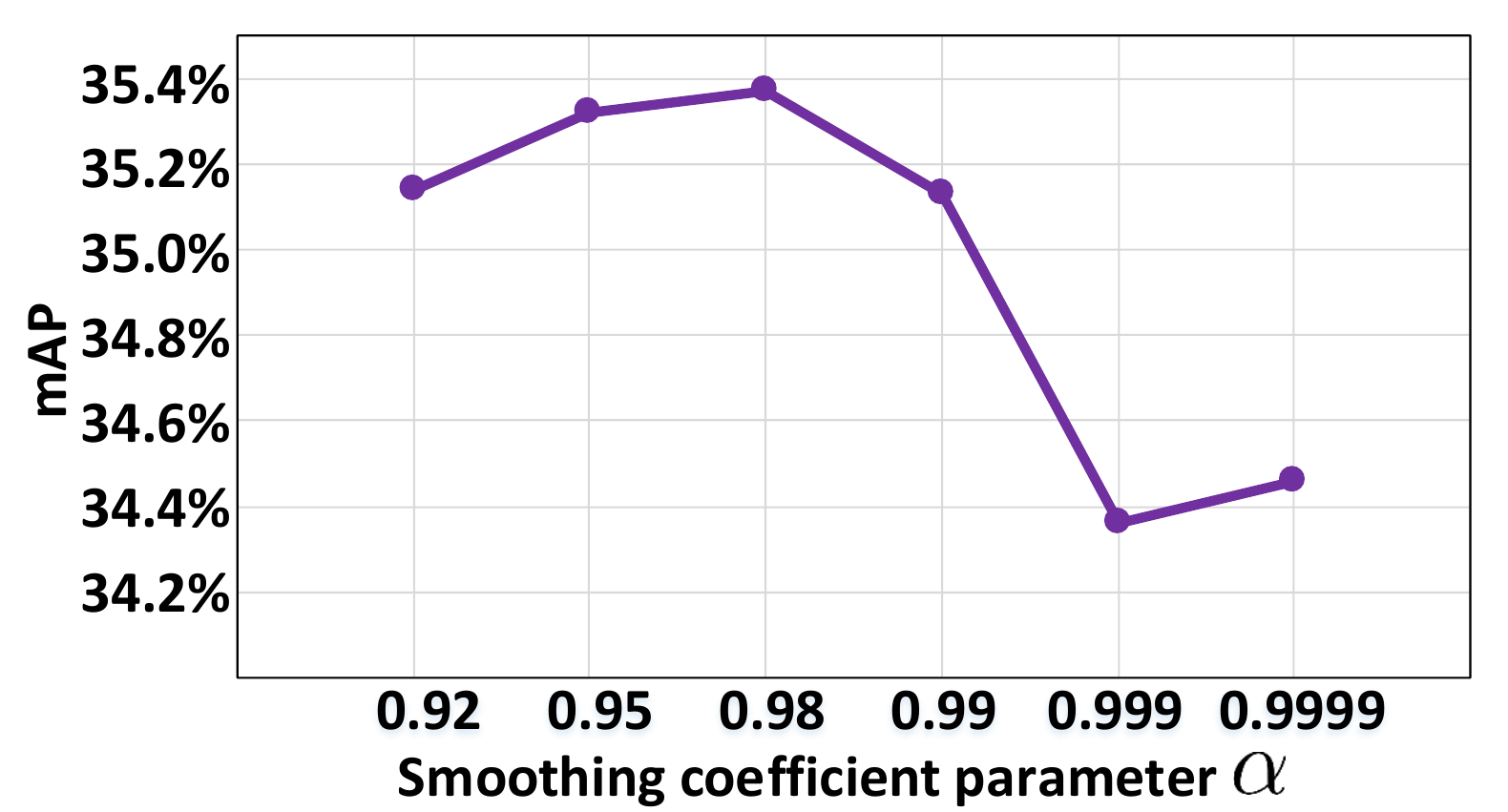}}
	\vspace{-0.05in}
	\caption{\small Effect of parameters $\lambda$ and $\alpha$ on \textbf{C} $\rightarrow$ \textbf{F} transfer.}
	\vspace{-0.15in}
	\label{fig:tradeoff}
\end{figure}

\textbf{Effect of the Parameters $\lambda$ and $\alpha$.}
To clarify the effect of tradeoff parameter $\lambda$ in Eq.(\ref{eq:OPT:Loss:eq-1}) and smoothing coefficient parameter $\alpha$ in Eq.(\ref{eq:OPT:Loss:eq-2}), we show the performance curves with different tradeoff/smoothing coefficient parameters in Figure \ref{fig:tradeoff}. As shown in the figure, we can see that both mAP curves of $\lambda$ and $\alpha$ are generally like the ``$\land$" shapes when $\lambda$ varies in a range from $0.1$ to $5.0$ and $\alpha$ varies in a range from $0.92$ to $0.9999$. The best performance is achieved when $\lambda$ is $1.0$ and $\alpha$ is about $0.98$.

\begin{figure}[!tb]
	\centering {\includegraphics[width=0.43\textwidth]{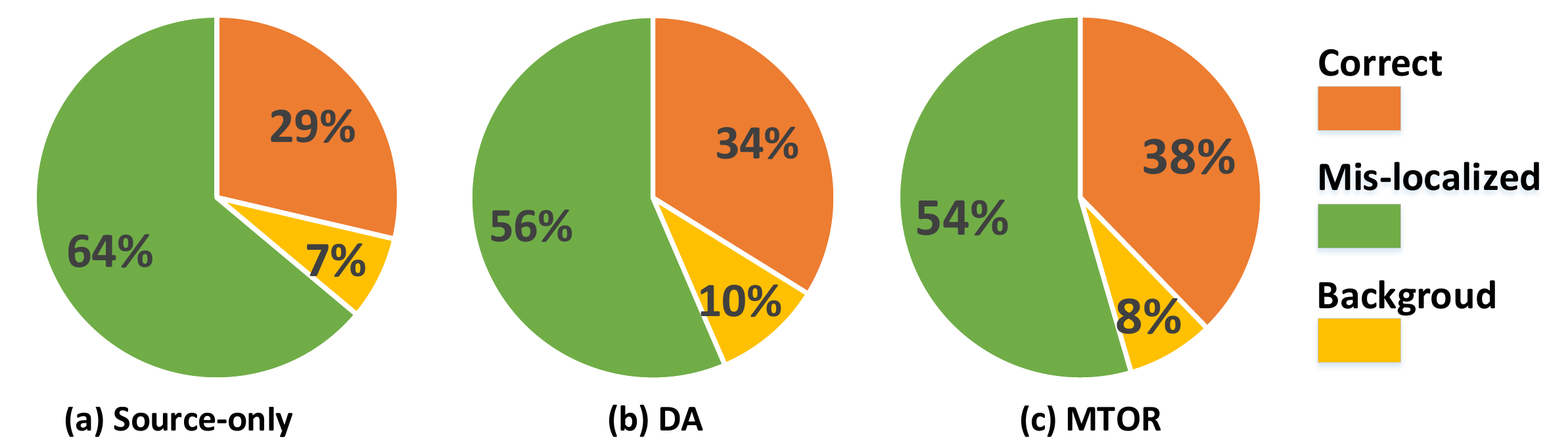}}
	\vspace{-0.05in}
	\caption{\small Error analysis of highest confident detections on \textbf{C} $\rightarrow$ \textbf{F}.}
	\vspace{-0.3in}
	\label{fig:pie}
\end{figure}

\textbf{Error Analysis of Highest Confident Detections.}~To further clarify the effect of the proposed region-level and graph-structured consistencies in Mean Teacher paradigm, we analyze the accuracies of Source-only, DA and MTOR caused by the highest confident detections on \textit{Foggy~Cityscapes} for \textbf{C} $\rightarrow$ \textbf{F} transfer. We follow \cite{chen2018domain,hoiem2012diagnosing} and categorize the detections into 3 types: \textbf{Correct} (IoU with ground-truth $\ge$ 0.5), \textbf{Mis-Localized} (0.5 $>$ IoU with ground-truth $\ge$ 0.3) and \textbf{Background} (IoU with ground-truth $<$ 0.3). For each class, we select top-$K$ predictions where $K$ is the number of ground-truth bounding boxes in this class. We report the mean percentage of each type across all categories in Figure \ref{fig:pie}. Compared to Source-only, DA and our MTOR clearly improve the number of correct detections (orange color) and reduce the number of false positives (other colors). Moreover, by leveraging region-level and graph-structured consistencies in Mean Teacher, MTOR leads to both smaller mis-localized and background errors than DA.

\begin{figure}[!tb]
	\centering {\includegraphics[width=0.32\textwidth]{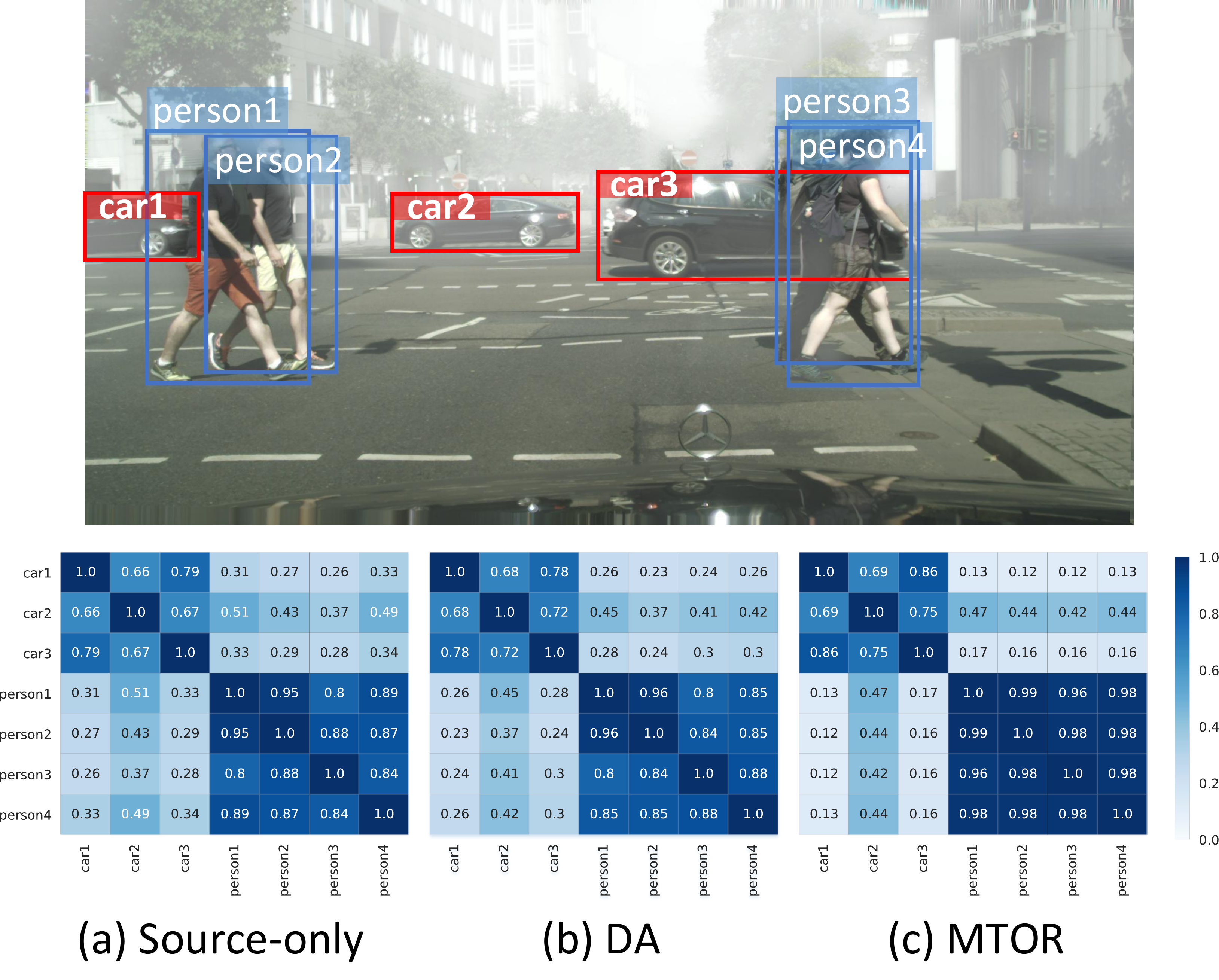}}
	\vspace{-0.1in}
	\caption{\small Visualization of relational graph on \textit{Foggy Cityscapes}.}
	\label{fig:graph}
	\vspace{-0.25in}
\end{figure}

\textbf{Visualization of Relational Graph.}
Figure \ref{fig:graph} further shows the visualization of an exemplar relational graph~(i.e., the affinity matrix) learned by Source-only, DA and MTOR on \textit{Foggy Cityscapes} for \textbf{C} $\rightarrow$ \textbf{F} transfer. For each approach, we extract the region representation of each ground-truth region and construct the relational graph by computing cosine similarity between every two regions. Note that the first three regions belong to \textit{car} class and the rest four regions fall into \textit{person} class. Thus we can clearly see that most intra-class similarities of MTOR are higher than those of Source-only and DA. The results demonstrate the advantage of enforcing intra-graph consistency in MTOR, leading to more discriminative region feature for object detection.

\vspace{-0.15in}

\section{Conclusions}\label{sec:CON}

We have presented Mean Teacher with Object Relations (MTOR), which explores domain adaptation for object detection in an unsupervised manner. Particularly, we study the problem from the viewpoint of both region-level and graph-structured consistencies in Mean Teacher paradigm. To verify our claim, we have built two relational graphs that capture similarities between pairs of regions for teacher and student respectively. The region-level consistency is to align the region-level predictions between teacher and student, which facilitates domain adaptation at region-level. The inter-graph consistency further matches the graph structures between teacher and student, pursuing a noise-resistant holistic graph structure on target domain. In addition, intra-graph consistency is utilized to enhance the similarity between regions of same class in student, which ideally leads to graph with lower intra-class variation. Experiments conducted on the transfers across \textit{Cityscapes}, \textit{Foggy Cityscapes}, and \textit{SIM10k} validate our proposal and analysis. More remarkably, we achieve state-of-the-art performance of single model on synthetic-to-real image transfer in \textit{Syn2Real} detection dataset.

\textbf{Acknowledgments.} This work was supported in part by National Key R\&D Program of China under contract No. 2017YFB1002203 and NSFC No. 61872329.

{\small
	\bibliographystyle{ieee}
	\bibliography{egbib}
}

\end{document}